\documentclass[runningheads]{llncs}
\usepackage{graphicx}
\usepackage{color,soul}
\usepackage{newtxtext,newtxmath}

\begin{document}
\title{Generating Responses Expressing Emotion in an Open-domain
Dialogue System}
%
%
\author{Chenyang Huang\orcidID{0000-0003-2811-6008} \and Osmar R. Za\"{\i}ane\orcidID{0000-0002-0060-5988}}
\authorrunning{C. Huang et al.}
%
\institute{ Department of Computing Science, \\
University of Alberta, Edmonton, Canada \\
 \email{\{chuang8,zaiane\}@ualberta.ca}}
\maketitle              
\begin{abstract}
Neural network-based Open-ended conversational agents automatically generate responses based on predictive models learned from a large number of pairs of utterances. The generated responses are typically acceptable as a sentence but are often dull, generic, and certainly devoid of any emotion. 
In this paper we present neural models that learn to express a given emotion in the generated response. We propose four models and evaluate them against 3 baselines. An encoder-decoder framework-based model with multiple attention layers provides the best overall performance in terms of expressing the required emotion.  While it does not outperform other models on all emotions, it presents promising results in most cases.

\keywords{Open-domain dialogue generation  \and Emotion \and Seq2seq \and Attention mechanism}

\end{abstract}


\section{Introduction}
\label{sec:intro}
Open-domain conversational systems \cite{bickmore2005establishing,ritter2011data,bessho2012dialog} tackle the problem of generating relevant responses given an utterance as input. 
Compared to the non-open-domain scenario, also known as task-oriented dialogue generation, where it is possible for agents to rely on knowledge for a narrowed domain and detect intent then use specific templates to generate responses, such as a travel booking system \cite{xu2000task}.

Open-domain
dialogue systems
have seen a growing interest in recent years thanks to
neural dialogue generation systems, based on deep learning models.
These systems do not encode dialog structure and are entirely data-driven. They
learn to predict the maximum-likelihood estimation (MLE) based on a large training corpus. The machine learning-based system basically learns to predict the words and the sentence to respond based on the previous utterances. However, while such a system can generate grammatically correct and human-like answers, the 
responses are often 
generic and non-committal instead of being specific and emotionally intelligent.

\begin{figure}
    \centering
    \includegraphics[width=0.7\textwidth]{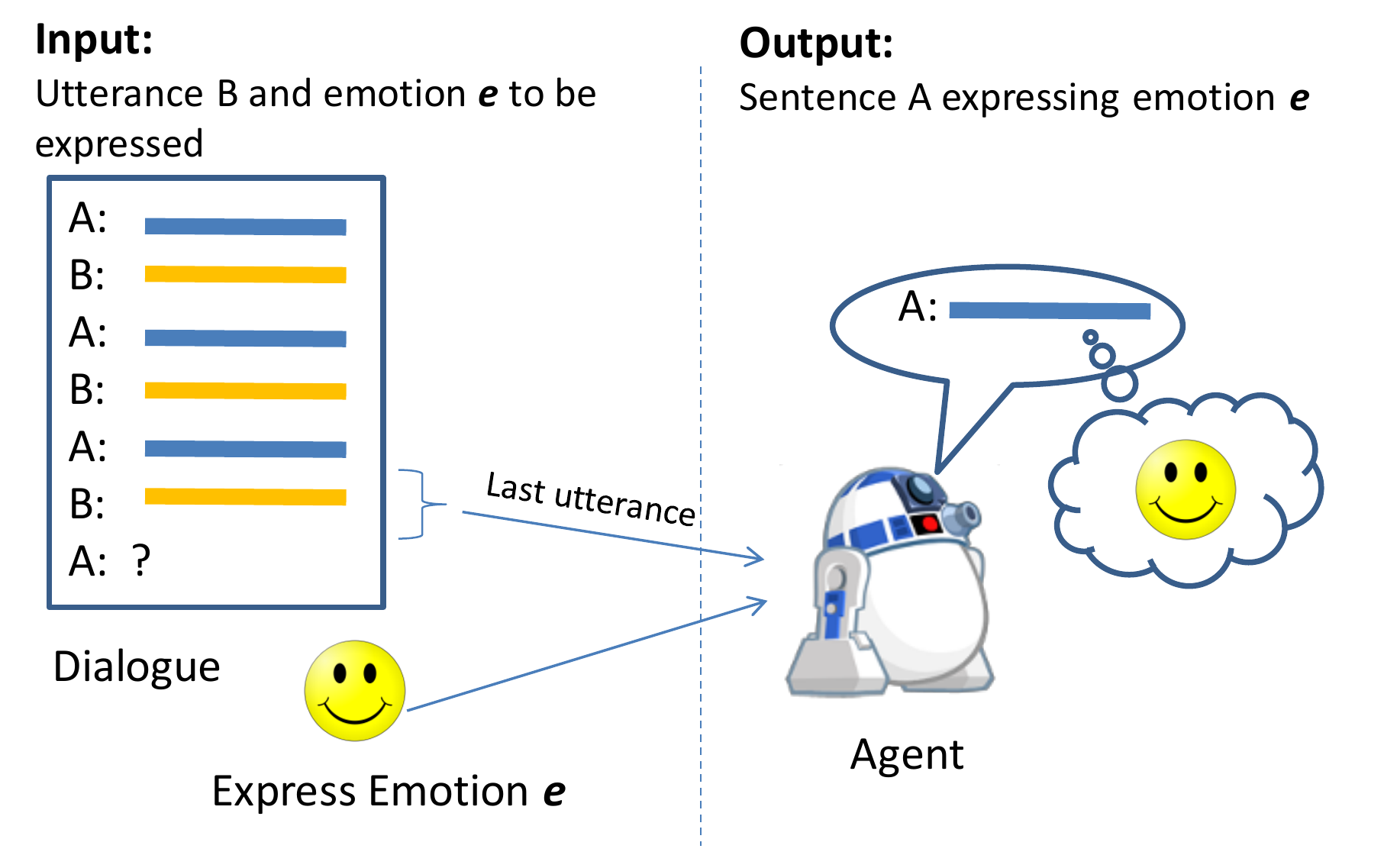}
    \caption{The task of generating response given emotions}
    \label{fig:task}
\end{figure}

However, an absolute ``automatic" system may find itself in situations where any inattentive response, even if correct and to the topic, may be improper or even negligent or offending. For example, if a person is expressing loneliness, or the death of a friend, the response should better be expressing empathy and support rather than a generic and careless possible response.  In this work, we are tackling the problem of how to control the emotions expressed in generated responses. 
As shown in Fig.~\ref{fig:task}, the proposes methods take as input a source sentence as well as an emotion to be expressed. It will not only respond with an relative sentence, but also express the given emotion in the response. 

To this end, a sufficiently large dialogue corpus labeled with emotions is required to train such a system which is both open-domain and emotionally intelligent. We follow the pipelines described in \cite{huang2018automatic}: 

\begin{enumerate}
    \item Train a classifier for emotion mining from text \cite{Shahraki17}.
    \item Label the Opensubtitles dataset \cite{lison2016opensubtitles2016} with emotions using the classifier. 
    \item Design and train models using the labeled dialogue corpus.
    \item Evaluate the models.
\end{enumerate}

In addition, we proposed 4 models: \textit{Dec-start}, \textit{Enc-att}, \textit{Dec-proj} and \textit{Dec-trans}. Experiment results show that all the proposed models are capable of the task and \textit{Enc-att} outperforms the baseline models in \cite{huang2018automatic} without adding too many parameters to the neural models.



\section{Related Work}
Emotion category classification serves as one of the basics of this work. The task of emotion mining from text is mainly composed of the following 3 aspects. 
\begin{itemize}
    \item Identify the categories.
    \item Obtain a large, high quality labeled dataset.
    \item Train a good classifier for emotions from text.
\end{itemize}
P. Ekman, one of the earliest emotion theorists, suggested 6 basic emotions in 1972 \cite{ekman1972emotion}: \textit{anger, disgust, fear, joy, sadness} and \textit{surprise}. The following work by P. Shaver \cite{shaver1987emotion} and W. G. Parrott \cite{parrott2001emotions} suggest removing \textit{disgust} and adding \textit{love}. A. G. Shahraki \cite{Shahraki17,Shahraki17b} combine the aforementioned emotion models by involving two additional emotions (\textit{guilt} and \textit{thankfulness}) but keeping \textit{disgust}.

Human annotation is one straightforward approach to obtain labeled datasets but it is costly. Only very small amount of manually labeled categorical dataset is available at the time of writing this paper. As an alternative, distant supervision \cite{mintz2009distant} has been investigated in many emotion detection researches \cite{mohammad2012emotional}
and proven to be efficient. Generally, they harvest tweets with emotion-carrying hashtags which are used as a surrogate of emotion labels. 

Having tweets labeled with emotions, training a classifier is a task of supervised text classification which has already been a well-studied area\cite{silva2011symbolic,lin2017structured}. The recent state-of-art models are usually neural network models\cite{zhou2016text} with pre-trained 


With the rise of deep learning, the success of the technology was also demonstrated in automatic response generation. The Sequence-to-sequence (Seq2seq) model which was shown effective in machine translation\cite{sutskever2014sequence}, was adopted in response generation for open domain dialogue systems  \cite{vinyals2015neural}. Instead of predicting a sequence of words in the target language from a sequence of words in the source language, the idea is to predict a sequence of words as a response of another sequence of words. In a nutshell,  Seq2seq models are a class of models that learn to generate a sequence of words given another sequence of words as input. Many works based on this framework have been conducted to improve the response quality from different points of view. 
Reinforcement learning has also been adopted to force the model to have longer discussions \cite{li2016deep}. \cite{serban2017hierarchical} proposed a hierarchical framework to process context more naturally. Moreover, there are also attempts to avoid generating dull, short responses \cite{li2017data,li2017adversarial}. 

The work in \cite{li2016persona} is able to generate personalized responses given a specific speaker, which can be considered as one of the first attempts that control the generations of seq2seq models. In terms of controlling emotions, \cite{ZhouHZZL18} tackles this problem with a sophisticated memory mechanism while \cite{huang2018automatic} uses three concise but efficient models to achieve equally good performance. 

\section{Seq2seq with Attention}
\label{sec:seq2seq}
Seq2Seq is a conditional language model which takes as input source-target pairs $(X, Y)$, where $X = {x_1, x_2, \cdots, x_m}$ and $Y = {y_1, y_2, \cdots, y_n}$ are sentences consisting of sequences of words. By maximizing the probability of $P(Y|X)$, these models can generate estimated sentences $\hat{Y}$ given any input $X$. 

Despite the variants of Seq2seq models, they usually consist of two major components: encoder and decoder. Such models can be referred as an encoder-decoder framework. The encoder will embed a source message into a dense vector representation $\boldsymbol{s}$ which is then fed into the decoder. The encoder and decoder are usually randomly initialized and jointly trained afterwards. 

The decoder will generate $\hat{Y} = \hat{y_1}, \hat{y_2}, \cdots$ in an autoregressive fashion. This procedure can be described as $\boldsymbol{s} = \textrm{Encoder}(X)$, $\hat{Y} = \textrm{Decoder}(Y, \boldsymbol{s})$. 

The choice of our encoder is an LSTM~\cite{hochreiter1997long} and it can be formulated as the following:

\begin{align}
\label{eq:1}
    h_t^{En}, c_t^{En} &= \text{LSTM}^{En}(M(x_i), [h_{t-1}^{En}; c_{t-1}^{En}]) \notag \\
    h_0^{En} &= c_0^{En} = \pmb{0}
\end{align}

Where $h_t^{En}$ and $c_t^{En}$ are encoder's hidden state and cell state at time $t$. $M(x)$ is the vector representation of word $x$ \cite{mikolov2013distributed}. In our experiments, we apply the state-of-the-art \textit{FastText} \cite{joulin2016fasttext} pre-trained model. 

Adapting attention mechanism in sequence generation has shown promising improvement \cite{bahdanau2014neural,luong2015effective}. In our case, we use the global attention with general score function \cite{luong2015effective} under the assumption that generated words can be aligned to any of the words in the previous dialogue utterance. We use another LSTM to decode the information. The decoder with attention can be described as:

\begin{align}
    \pmb{h}^{En} &= [h_1^{En}, h_2^{En}, \cdots, h_m^{En}]  \\
    \alpha_t &= \text{softmax}(h_t^{De} \text{Tanh}(W_a \pmb{h}^{En}))  \label{eq3} \\
    \hat{h}_{t} &= \alpha_t \cdot \pmb{h}^{En} \label{eq4}  \\  
    h_t^{De}, c_t^{De} &= \text{LSTM}^{De}\Big(M(y_i), \big[\hat{h}_{t-1}; c_{t-1}^{De}\big]\Big) \label{eq5} \\ 
    y_i &=  \text{argmax}\Big(\text{softmax}\big(Proj(h_i^{De}) \big) \Big) \label{eq6} \\ 
    \hat{h}_{0} &= h^{En}_m , ~ c_0^{De} = c^{En}_m  \nonumber
\end{align}

Where $h_t^{De}$ and $c_t^{De}$ are hidden state and cell state. $\alpha_t$ is the attention weights over all hidden states of encoder. $W_a$ is a trainable matrix which is initialized randomly.


The seq2seq with attention mechanism is shown in Fig.~\ref{fig:compare}, where \textbf{E} and \textbf{D} represent two LSTM models for encoder and decoder respectively. $\alpha_t$ in Equation~\ref{eq3} is known as attention scores. According to equations~(\ref{eq3}),(\ref{eq4}) and (\ref{eq5}), the attention score has to be calculated at every decoding step repeatedly. In Fig.~\ref{fig:compare}, the illustration of the attention layer is only for $h_1^{De} \,\to\, h_2^{De}$. ${Proj(h_i^{De})}$ is a linear layer that projects the hidden state of time step $i$ to the one dimensional space of vocabulary. After normalization, e.g. softmax, the output of ${Proj(h_i^{De})}$ is the probability of the next token. The most possible one is considered as $\hat{y_i}$. During the training, the loss is calculated by comparing the difference between $Y$ and $\hat{Y}$. The resulting cross entropy loss can be calculated by Equation~\ref{eq:ce_loss},

\begin{equation} \label{eq:ce_loss}
   H(Y, \hat{Y}) = \sum _{i=1}^{n}\ {\bigg [}y_{i}\log {\hat {y}}_{i}+(1-y_{i})\log(1-{\hat {y}}_{i}){\bigg ]} 
\end{equation}

\section{Emotion Injection}

As mentioned above in Section \ref{sec:seq2seq}, general seq2seq models are learning the probability of $P(Y|X)$. While in the task of controlling responses by an instructed emotion, each $(X, Y)$ pair is assigned with an additional desired response emotion $e$. 
The goal is therefore to estimate the target sequence $Y$ given the joint probability of $X \cap e$,  which can be written as $P(Y|X \cap e)$,  

\subsection{Baseline models}

We propose three models (\textit{Enc-bef}, \textit{Enc-aft} and \textit{Dec}) in \cite{huang2018automatic}. \textit{Enc-bef} and \textit{Enc-aft} are models that inject an emotion $e$ in the encoder by putting special tokens before or after the input sequence $X$. The $Dec$ model, on the other hand, puts $e$ at each decoding step, which is similar to the method in \cite{li2016persona}. $Dec$ changes (\ref{eq5}) to the following:

\begin{equation}
\label{eq:dec_rep}
    h_t^{De}, c_t^{De} = \text{LSTM}^{De}\Big(M(y_i), [\hat{h}_{t-1}; c_{t-1}^{De}; v_e\big]\Big)
\end{equation}

In Equation~(\ref{eq:dec_rep}) $v_e$ is randomly initialized and trainable vector for each of the emotions. To be more precise, we will refer to the $Dec$ model as \textit{Dec-rep} in the follow text.

\subsection{Proposed models}
The Encoder-Decoder framework provides us with a flexible foundation which makes joining additional modules into it straightforward and intuitive. In this work, inspired by \cite{huang2018automatic} and \cite{ZhouHZZL18}, four models are proposed.  

\subsubsection{Dec-start}
In \cite{huang2018automatic} \textit{Enc-bef} and \textit{Enc-aft} models have been shown to be successful and effective. By creating a special token $T_e$ for every emotion, these two methods are essentially modifying $X$ to $\big[T_e; X\big]$ or $\big[X; T_e\big]$ in both training and evaluating. As shown in Fig.~\ref{fig:compare}, to start decoding, a special token \textit{<s>} is fed into the decoder. $h_1^{De}$ is obtained by calculating $LSTM^{De}(M(\text{<s>}),[h_m^{En}, c_m^{En}])$. 

In this Dec-start model, we simply substitute the start token \textit{<s>} with an emotion token $T_e$ as shown in Equation~(\ref{eq:dec_start}). 

\begin{equation}
\label{eq:dec_start}
    h_1^{De},c_1^{De} = LSTM^{De}\Big(M(T_e),\big[h_m^{En}, c_m^{En}\big]\Big)
\end{equation}

\subsubsection{Dec-trans}
 
As an alternative, we multiply the $h_t^{De}$ with another matrix to transform the hidden state of time $t$ with respect to the emotion to be expressed. Denote the transforming matrix as $Trans_e$, Equation~\ref{eq6} is changed as following:

\begin{equation}
    y_i =  \text{argmax}\bigg(\text{softmax}\Big(Proj \big(Trans_e(h_i^{De}) \big) \Big) \bigg)  
\end{equation}

\subsubsection{Dec-proj}
\cite{ZhouHZZL18} also proposed an external memory which maps the hidden state $h_t^{De}$ into a slightly different vocabulary space for each of the emotions. By taking a step forward, we propose a \textit{Dec-proj} model which will make $h_t^{De}$ to totally independent vocabulary spaces. This is done by making unique projection layer $Proj_e$ for each of the emotion. Equation~\ref{eq6} is thus changed to the following:

\begin{equation}
   y_i =  \text{argmax}\Big(\text{softmax}\big(Proj_e(h_i^{De}) \big) \Big)
\end{equation}

\subsubsection{Enc-att}
The attention mechanism has been proven to be very powerful in many sequence to sequence tasks.  \cite{vaswani2017attention} even outperformed many tasks by using an attention only encoder-decoder model. \cite{luong2015effective} proposed three methods to calculate the attention score. The one we chose in Equation~\ref{eq3} is referred to as \textit{general} score in their paper. It is a parameterized method compared 
to \textit{dot} score.  Since the general attention has individual parameters, making different attention layer for different emotion is possible as well. The \textit{Enc-att} model changes Equation~\ref{eq3} to the following:

\begin{equation}
    \alpha_t = \text{softmax}(h_t^{De} \text{Tanh}(W_e \pmb{h}^{En})) 
\end{equation}

\begin{figure}[ht]
    \centering
    \includegraphics[width=0.8\textwidth]{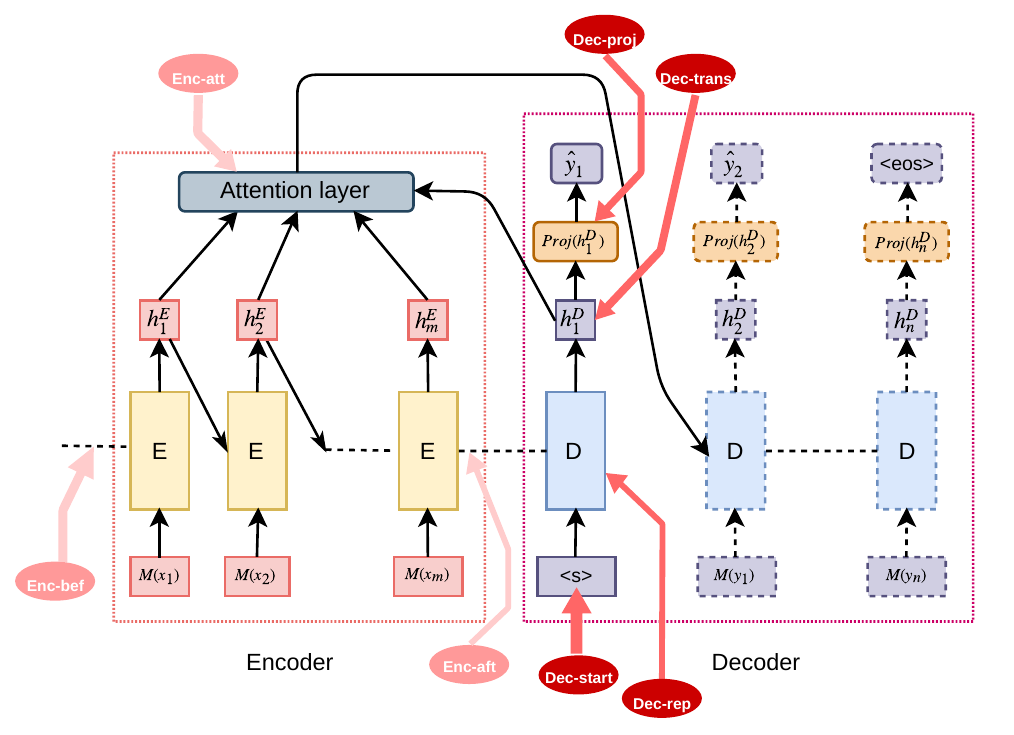}
    \caption{Comparison among the four proposed models and three baseline models} 
    \label{fig:compare}
\end{figure}

Compared to the original equation, the universe matrix for calculating attention score $W_a$ is replace with $W_e$ for each of the given emotion $e$. 
Fig.~\ref{fig:compare} shows where the emotion is injected into standard   seq2seq with attention model. 

\section{Dataset}

As mentioned in Section\ref{sec:intro}, there is no dataset that contains pairs of dialogue exchanges and corresponding emotions. But there are categorical datasets for emotion classification in text. For example, \textit{Cleaned Balanced Emotional Tweets (CBET)} dataset \cite{Shahraki17,Shahraki17b}, \textit{Twitter Emotion Corpus (TEC)} \cite{mohammad2012emotional} and the \textit{International Survey on Emotion Antecedents and Reactions (ISEAR) } \cite{scherer1994evidence}. There are some other dimensional datasets but they can not directly fit into this task. Table~\ref{table:dataset} shows more details about the aforementioned datasets.

\begin{table}[ht]
\centering
\caption{Details of three categorical datasets for emotion classification}
    \begin{tabular}{c|c|c|c}
    \hline \hline 
    Dataset & \# of categories & \# instances & Emotions   \\  \hline                                   
    CBET    & 9                & 81,163        & anger, surprise, joy, love, sadness, fear, disgust, \\
            &                 &               & guilt , thankfulness \\  \hline 
    TEC     & 6                & 21,051       & anger, disgust, fear, joy, sadness, surprise \\  \hline 
    ISEAR   & 7                & 7,666         & joy, fear, anger, sadness, disgust, shame, guilt  \\  \hline \hline 
    \end{tabular}
\label{table:dataset}
\end{table}

Considering the size of the datasets and also for consistency with the work in \cite{huang2018automatic}, we choose the CBET dataset to train a text classifier and use that to tag the corpus which is used to train a dialogue model.

By applying a bidirectional LSTM \cite{graves2005bidirectional} model with self-attention \cite{lin2017structured} We achieve a slightly better results than that in \cite{huang2018automatic}, which has a F1-score of 54.33\% with precision of 66.20\% and recall of 51.29\%.

The OpenSubtitles dataset \cite{lison2016opensubtitles2016} is one of the largest and most popular dataset to train open domain dialogue systems. Following the work in \cite{huang2018automatic}, we use the pre-processed data by \cite{li2016persona} and further remove duplicate lines. The total number of utterances is 11.3 million, each utterance has at least 6 words.

In addition, a threshold is applied to approximate a \textit{Non-emotion} category.  This means, in evaluation, if the most possible emotion of an instance still has a very low `confidence', this instance would be considered as not containing any of those emotions. In our experiment, by setting the threshold to 0.35, approximately 35\% of the sentences are below the threshold. \textit{Non-emotion} is treated as a special emotion when training the dialogue models, but it is not considered in the evaluation.
 
\section{Experiments}

\subsection{Parameters setup}
For the purpose of comparison, the parameters of the proposed models are set to as close to the baseline models as possible.  The dimension for both encoder LSTM and decoder LSTM is 600. The dropout ratio is 0.75. The choice of optimizer is Adam  \cite{kingma2014adam} with learning rate set to 0.0001. The number of the vocabulary is set to 25,000. \textit{FastText} \cite{joulin2016fasttext} pre-trained word embedding model is used and set to trainable. The size of held-out test set is 50k samples. The training and development split ratio is 0.95 to 0.05. The padding length is 30.

\subsection{Evaluation metric}

The main goal of this research is to check the ability of generating responses while given a specific emotion. That being said, the quality and relevance of generated responses are not the focus of this research.
Hence, our interest lies in checking if the generated sentences contains the instructed emotions or not. 
The size of the test set is 50k and 7 models are evaluated: 3 baselines and our 4 proposed models. Further more,  every instance in the test set is assigned to 9 emotions for the model. Thus, a total of 3,150k ($50k \times 7 \times 9$) responses are generated for evaluation. 

Fortunately, unlike the work by \cite{li2016persona}, expensive human evaluation is not needed. Instead, we evaluate the output using the emotion mining classifier again. Since every source sentence we generate 9 responses (i.e. one for each emotion), for each emotion category, we check the proportion of the responses where the corresponding emotions are indeed expressed. Such proportion is considered as the \textit{estimated accuracy}. Therefore, for each model, we can obtain 9 \textit{estimated accuracy} scores for the 9 emotions.


\section{Results and Discussion}

\subsection{Result analysis}
The \textit{estimated accuracy} scores of the 7 models are shown in Table~\ref{table:results}. Moreover, we draw the confusion matrices of the 4 proposed models to show the misclassification errors (Fig. \ref{fig:compare}). For the confusion matrices of the 3 baseline models, please refer to \cite{huang2018automatic}.
\begin{table}[h!] 
\centering
\caption{Per class accuracy of generated response}
\begin{tabular}{l|c|c|c|c|c|c|c}
\hline \hline
Emotion      & \textit{Enc-bef} & \textit{Enc-aft} & \textit{Dec-rep}   & \textit{\textbf{Dec-start}} & \textit{\textbf{Dec-trans}} & \textit{\textbf{Dec-proj}} & \textit{\textbf{Enc-att}} \\ \hline 
\textcolor{red}{anger}		& 60.18\%	& 62.30\% 	& 67.95\% 	& 66.81\% 	& 64.27\% 	& \textbf{78.48}\% 	& 65.09\% \\
\textcolor{red}{disgust} 	& 77.98\%	& 76.79\% 	& 79.02\% 	& 78.42\% 	& 78.33\% 	& \textbf{86.43}\% 	& 78.29\% \\
\textcolor{red}{fear} 		&\textbf{ 86.40}\% 	& 84.17\% 	& 83.52\% 	& 84.10\% 	& 77.15\% 	& 73.70\% 	& 86.00\% \\
\textcolor{red}{joy} 		& 45.69\%	& 41.15\% 	& 48.30\% 	& 47.42\% 	& 49.69\% 	& \textbf{59.12}\% 	& 38.71\% \\
\textcolor{red}{sadness} 	& 94.19\% 	& 93.98\% 	& 94.21\% 	& 94.18\% 	& 88.42\% 	& 89.83\% 	& \textbf{95.09}\% \\
\textcolor{red}{surprise} 	& 84.47\% 	& 85.09\% 	& 87.21\% 	& 80.55\% 	& 83.61\% 	& 80.56\% 	& \textbf{92.5}4\% \\
\textcolor{blue}{love} 		& 56.38\% 	& 54.69\% 	& 58.32\% 	& 54.25\% 	& 62.82\% 	& \textbf{85.14}\% 	& 64.56\% \\
\textcolor{blue}{thankfulness} & 87.69\% 	& 89.31\% 	& \textbf{90.83}\% 	& 89.44\% 	& 82.03\% 	& 61.80\% 	& 89.11\% \\
\textcolor{blue}{guilt} 		& 93.19\% 	& 92.17\% 	& 91.20\% 	& 90.68\% 	& 86.64\% 	& 50.92\% 	& \textbf{94.40}\% \\ \hline 
Average 	& 76.24\% 	& 75.52\% 	& 77.84\% 	& 76.21\% 	& 74.77\% 	& 74.00\% 	& \textbf{78.20}\% \\ \hline \hline 
\end{tabular}
\label{table:results}
\end{table}

From the table, we can see that despite the fact that the \textit{Enc-att} model only achieves 38.71\% accuracy for the emotion \textit{joy}, it still outperforms the others on most of the emotions. From Fig.~\ref{fig:compare}, one can observe a significant mismatch between \textit{joy} and \textit{thankfulness}. Instead of expressing \textit{joy}, \textit{Enc-att} conveys \textit{thankfulness} which could also be considered reasonable. However, \textit{love} is also confused with \textit{guilt}. Note that the measured accuracy is also subject to the accuracy of the emotion tagger used.

It is also noticeable that the performance of model \textit{Dec-start} is close to that of models \textit{Enc-bef} and \textit{Enc-aft}. This is expected considering the models are simply injecting the information of emotions by only one special token. The highlighted numbers in Table~\ref{table:results} show the best accuracy of each emotion. Model \textit{Dec-rep}, \textit{Dec-proj} and \textit{Enc-att} have at least 2 best scores whereas the others almost have none. 

\begin{figure*}[!ht]
    \centering
    \begin{tabular}{c @{\qquad} c }
        \includegraphics[width=.46\linewidth]{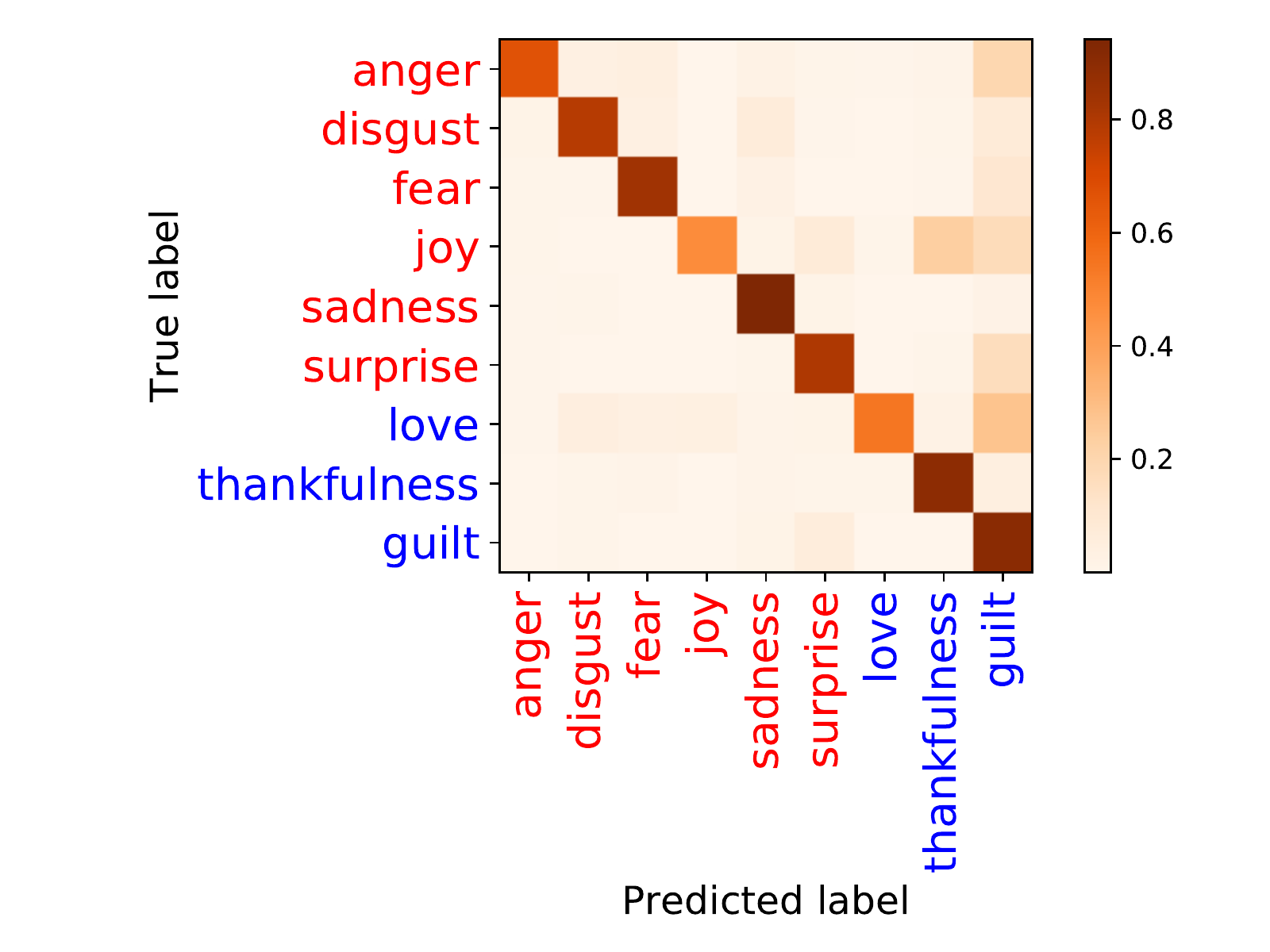} &
        \includegraphics[width=.46\linewidth]{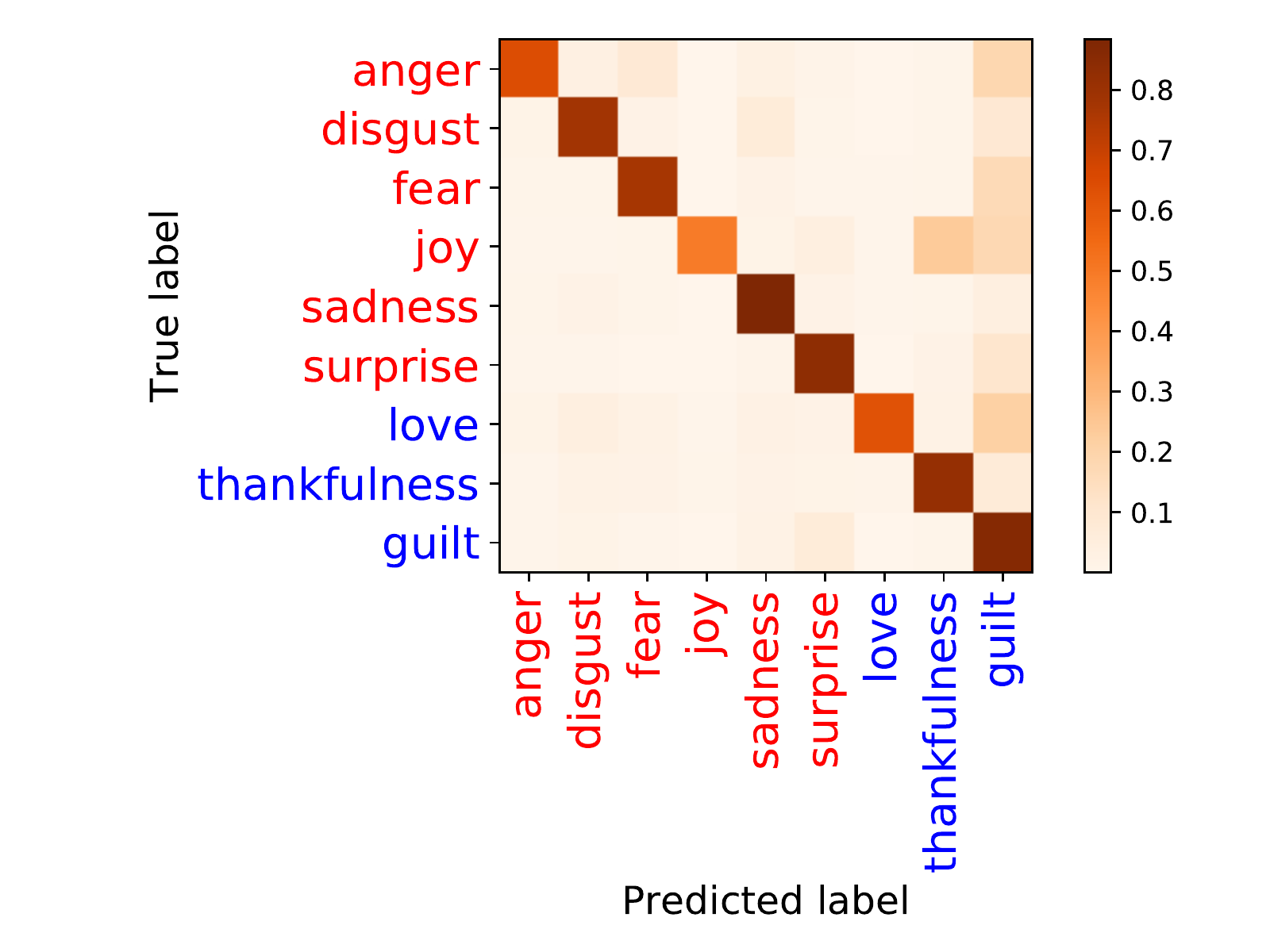} \\
        \small (a) \textit{Dec-start} & \small (b) \textit{Dec-trans} \\
        \includegraphics[width=.46\linewidth]{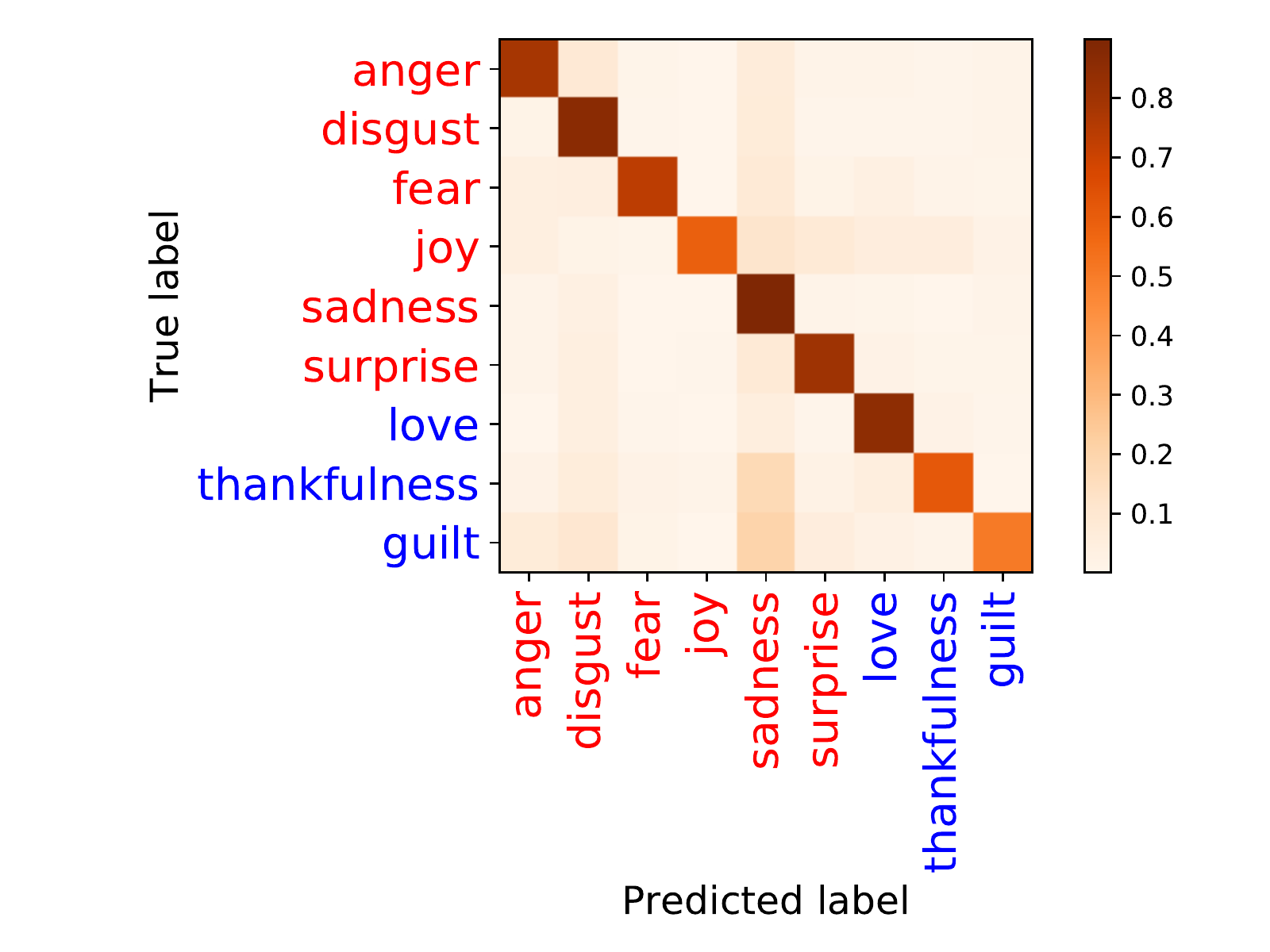} &
        \includegraphics[width=.46\linewidth]{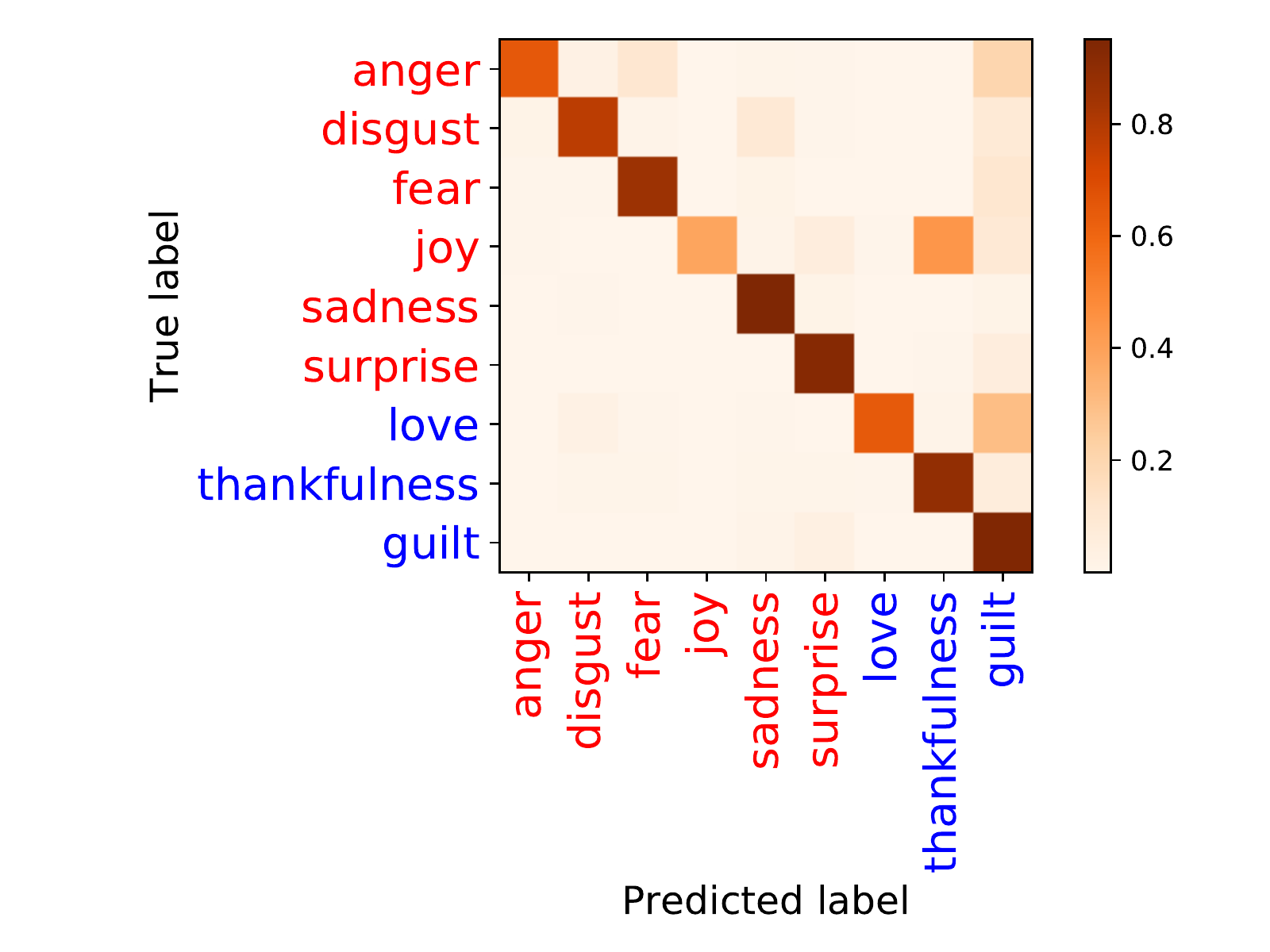} \\
        \small (c) \textit{Dec-proj} & \small (d) \textit{Enc-att}
    \end{tabular}
    \caption{Confusion matrices of 4 proposed models}
    \label{fig:confusion}
\end{figure*}

To compare the extended emotion model with Ekman's basic emotions, we highlight the two group of emotions in both Fig.~\ref{fig:confusion} and Table~\ref{table:results}. The emotions in red are the six basic emotions, the blue ones are those added by \cite{Shahraki17}.

\subsection{Enc-att model visualization}

\begin{figure*}[!ht]
    \centering
    \begin{tabular}{c @{\qquad} c }
    \includegraphics[width=.46\linewidth]{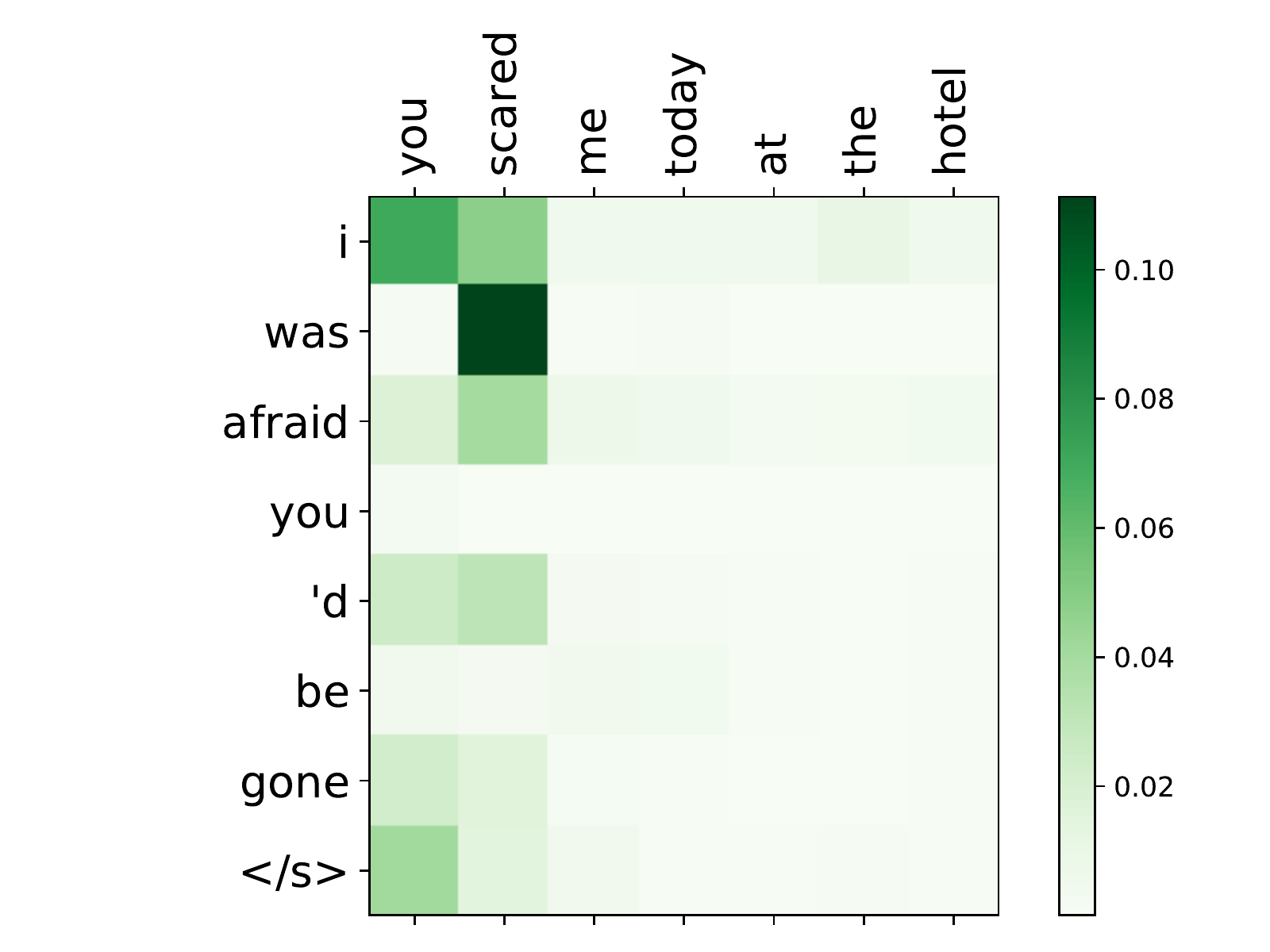} &
    \includegraphics[width=.46\linewidth]{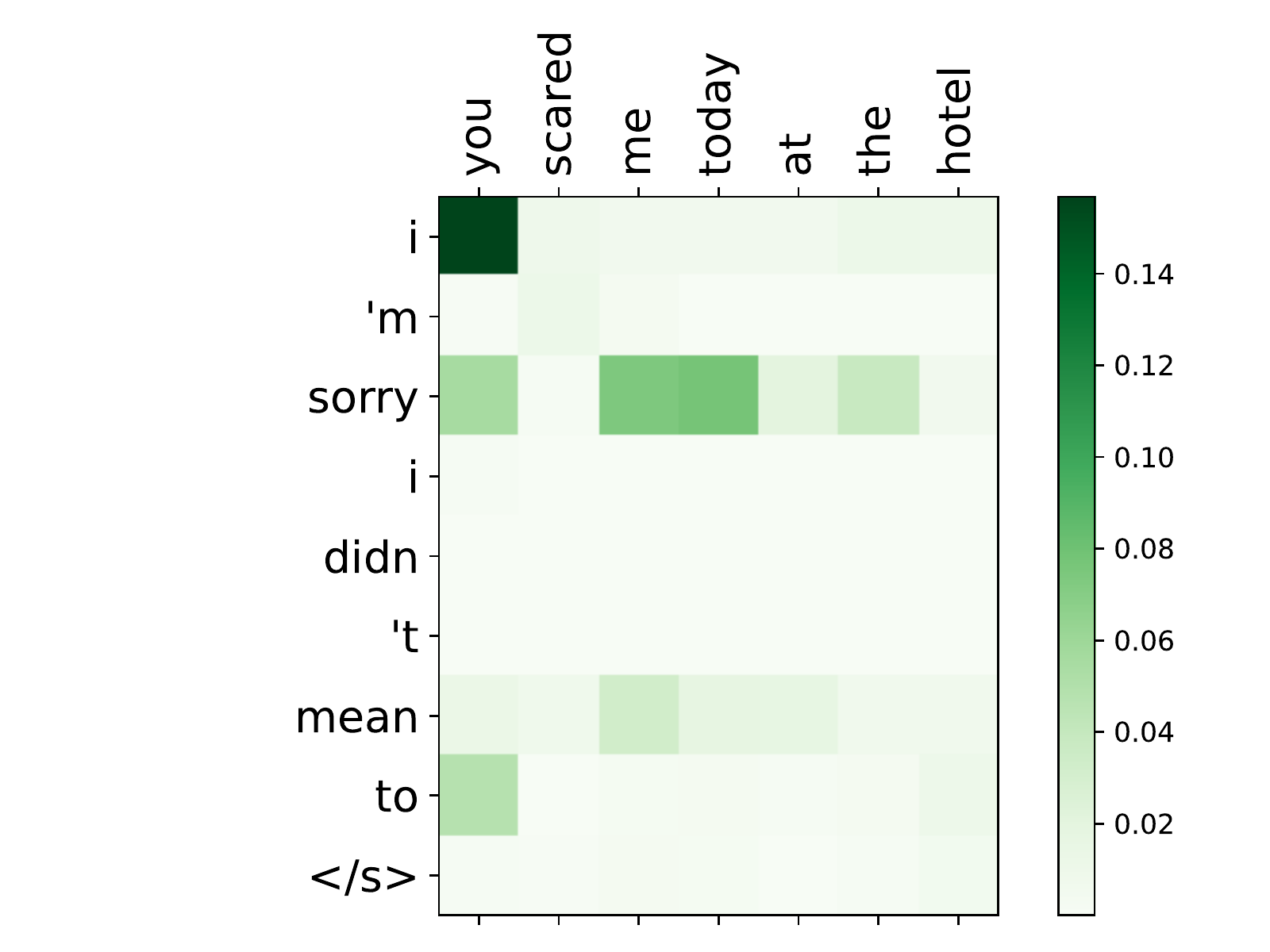} \\
    \small (a) fear & \small (b) guilt \\
    \includegraphics[width=.46\linewidth]{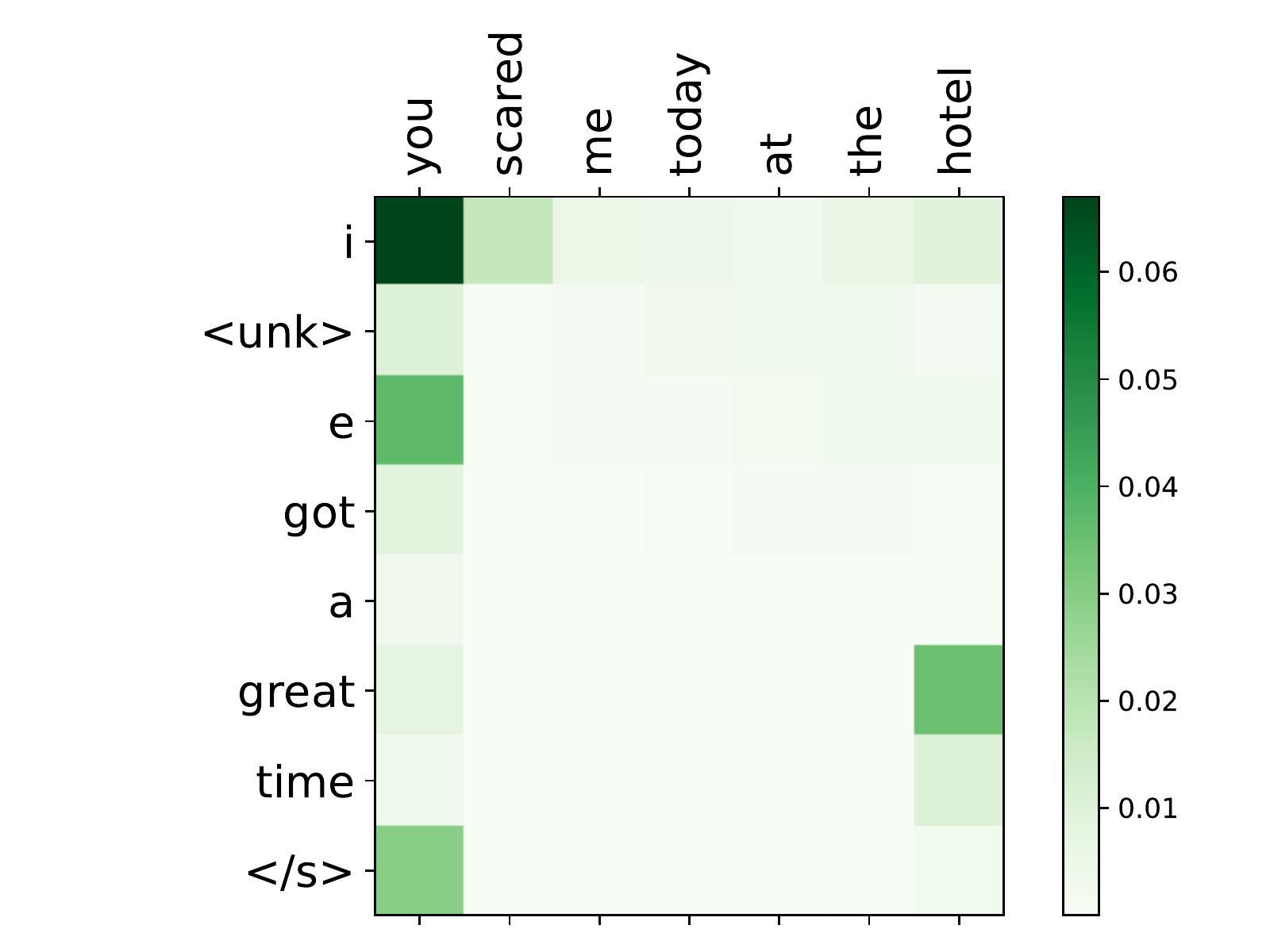} &
    \includegraphics[width=.46\linewidth]{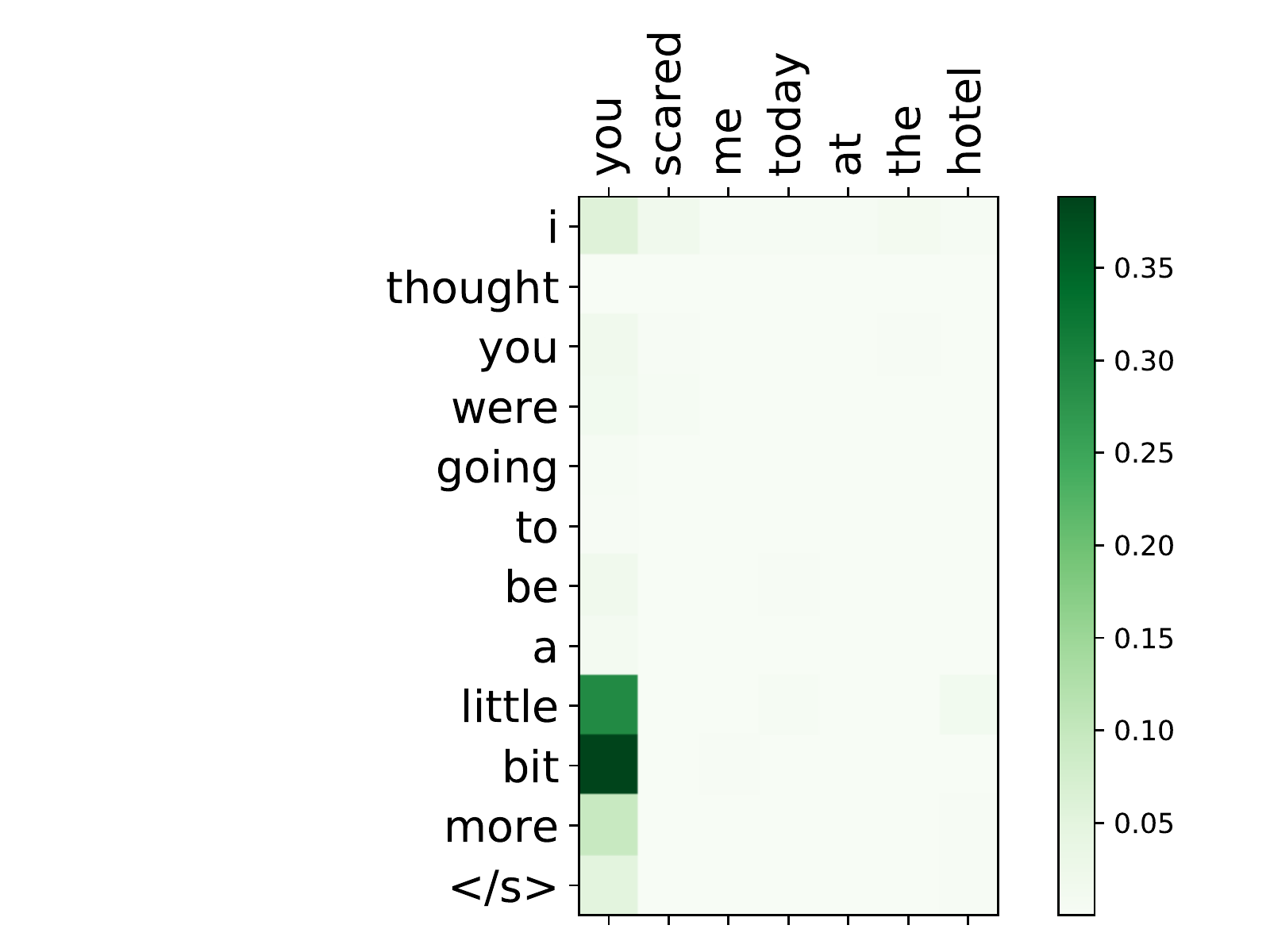} \\
    \small (c) joy & \small (d) sadness
    \end{tabular}
    \caption{An example of the attention scores of \textit{Enc-att} model}
    \label{fig:att}
\end{figure*}

To show how the \textit{Enc-att} model works, we chose an example utterance and show how the attention scores vary with respect to responses with different emotions. The attention is visualized by heatmaps in Fig.~\ref{fig:att}. To respond to the utterance \textit{You scared me today at the hotel}, the model focused on \textit{scared} when expressing \textit{fear}. When conveying \textit{guilt}, except for focusing on the pronoun \textit{you}, it focused on \textit{me} and \textit{today} and show a strong preference to using the word \textit{sorry}. When responding with \textit{joy}, it focused on the word \textit{hotel}. Interestingly, to response to the utterance with \textit{sadness}, the model did no pay attention to any words except for the pronoun, but it did try to answer with the phrase \textit{little bit more}.

\subsection{Parameter cost}
Apart from the performance of the models, another important comparison of deep learning models is their sizes. Considering that all the 7 models are based on the basic seq2seq with attention model. We only need to compare the additional parameters that are needed. Let's denote the size of vocabulary space as $|V|$, the length of source sentences as $m$, the dimension of the decoder LSTM as $D$, and the number of emotions as $S$. The comparison of the models in terms of these parameters is shown in Table~\ref{table:para}. It has to be mentioned that $S$ in our experiments is 10: 9 emotions plus an \textit{non-emotion} category.

\begin{table}[h!]
\centering
\caption{Comparison of the models in terms of additional space for required parameters}
\begin{tabular}{c|c|c}
\hline \hline
Model &  Additional para. in symbols & Additoinal para. in our exp. \\\hline 
Enc-bef     & 0                                  &    0          \\
Enc-aft     &  0                                 &    0          \\
Dec-rep     &      $D \times S $                 &   6,000             \\
Dec-start     &    0                             &    0         \\
Dec-trans     &     $D \times D \times S $       &      3,600,000          \\
Dec-proj     &     $|V| \times D \times S$      &      150,000,000            \\
Enc-att      &    $m \times D \times S$        &    180,000      \\ \hline \hline 
\end{tabular}
\label{table:para}
\end{table}

From the above table, \textit{Dec-proj} is the least cost efficient model. \textit{Dec-rep} and \textit{Enc-att} are both outperforming models considering their performance.  


\section{Conclusion and Perspectives}
In this work, we propose four models that are able to automatically generate a response while conveying a given emotion. We compare our models with the baseline models in \cite{huang2018automatic} in terms of both performance and efficiency. Our \textit{Enc-att} model outperforms the strongest baseline and we show how it works using attention heatmaps. \textit{Dec-rep} and \textit{Enc-att} turn out to be both effective and efficient.

However, in this work, we did not experiment with any combinations of the models. It is shown in \cite{ZhouHZZL18} that the combination of external and internal memory outperforms each of the single model. We think the combination of \textit{Dec-rep} and \textit{Enc-att} has a potential to give a better result. 
One major limitation of this work is that we heavily rely on the accuracy of the emotion mining classifier and assume it is of acceptable accuracy. 
Moreover, the main effort of this research lies on generating responses accurately and efficiently but without focusing on properties like grammar, relevance and diversity. 


%
%
%
\bibliographystyle{splncs04}
\bibliography{refs}

\end{document}